\pdfoutput=1

\documentclass[11pt]{article}

\usepackage{EMNLP2023}
\newcommand{\model}{\textsf{CoRelation}}
\usepackage{amsmath}
\usepackage{bbding}
\usepackage{times}
\usepackage{tablefootnote}
\usepackage{amsfonts}
\usepackage{latexsym}
\usepackage{multirow}
\usepackage[ruled,vlined]{algorithm2e}
\usepackage{url}
\usepackage{algpseudocode}
\usepackage{graphicx}
\usepackage{makecell}

\usepackage{colortbl}

\usepackage[T1]{fontenc}

\usepackage[utf8]{inputenc}

\usepackage{microtype}

\usepackage{inconsolata}

\title{\model: Boosting Automatic ICD Coding Through Contextualized Code Relation Learning}

\author{Junyu Luo$^1$, Xiaochen Wang$^1$, Jiaqi Wang$^1$, Aofei Chang$^1$, Yaqing Wang$^2$, Fenglong Ma$^1$\\
$^1$Pennsylvania State University, $^2$Google\\
  $^1$\texttt{\{junyu, xcwang, jqwang, aofei, fenglong\}@psu.edu}, $^2$\texttt{yaqingwang@google.com}
}

\begin{document}
\maketitle
\begin{abstract}
Automatic International Classification of Diseases (ICD) coding plays a crucial role in the extraction of relevant information from clinical notes for proper recording and billing. One of the most important directions for boosting the performance of automatic ICD coding is modeling ICD code relations. However, current methods insufficiently model the intricate relationships among ICD codes and often overlook the importance of context in clinical notes. In this paper, we propose a novel approach, a contextualized and flexible framework, to enhance the learning of ICD code representations. Our approach, unlike existing methods, employs a dependent learning paradigm that considers the context of clinical notes in modeling all possible code relations. We evaluate our approach on six public ICD coding datasets and the experimental results demonstrate the effectiveness of our approach compared to state-of-the-art baselines.
\end{abstract}
\section{Introduction}\label{sec:intro}
The International Classification of Diseases (ICD) is a standard code system devised by the World Health Organization (WHO), which has gained widespread adoption in electronic health records (EHR) and health insurance systems. 
Currently, medical coders review medical records and other relevant documents to extract information about the patient's conditions, the services provided, and any procedures performed. They then translate this information into standardized ICD codes. This procedure is usually called \emph{ICD coding}. 
However, manually assigning ICD codes is not only time-consuming but also error-prone. 


To alleviate this issue, the concept of \textbf{automatic ICD coding} has been proposed recently, which is viewed as a multi-label text classification task~\cite{mullenbach2018explainable}. The model is required to predict the probability distribution of ICD codes based on clinical notes, such as discharge summaries. Although various approaches ~\cite{xie2019ehr, li2020icd, cao2020hypercore, vu2020label, yuan2022code, yang2022knowledge} have been proposed to enhance the performance of automatic ICD coding, they still have several limitations in \emph{modeling relationships among ICD codes}.

\begin{figure*}[h]
\centering
\includegraphics[width=0.95\textwidth]{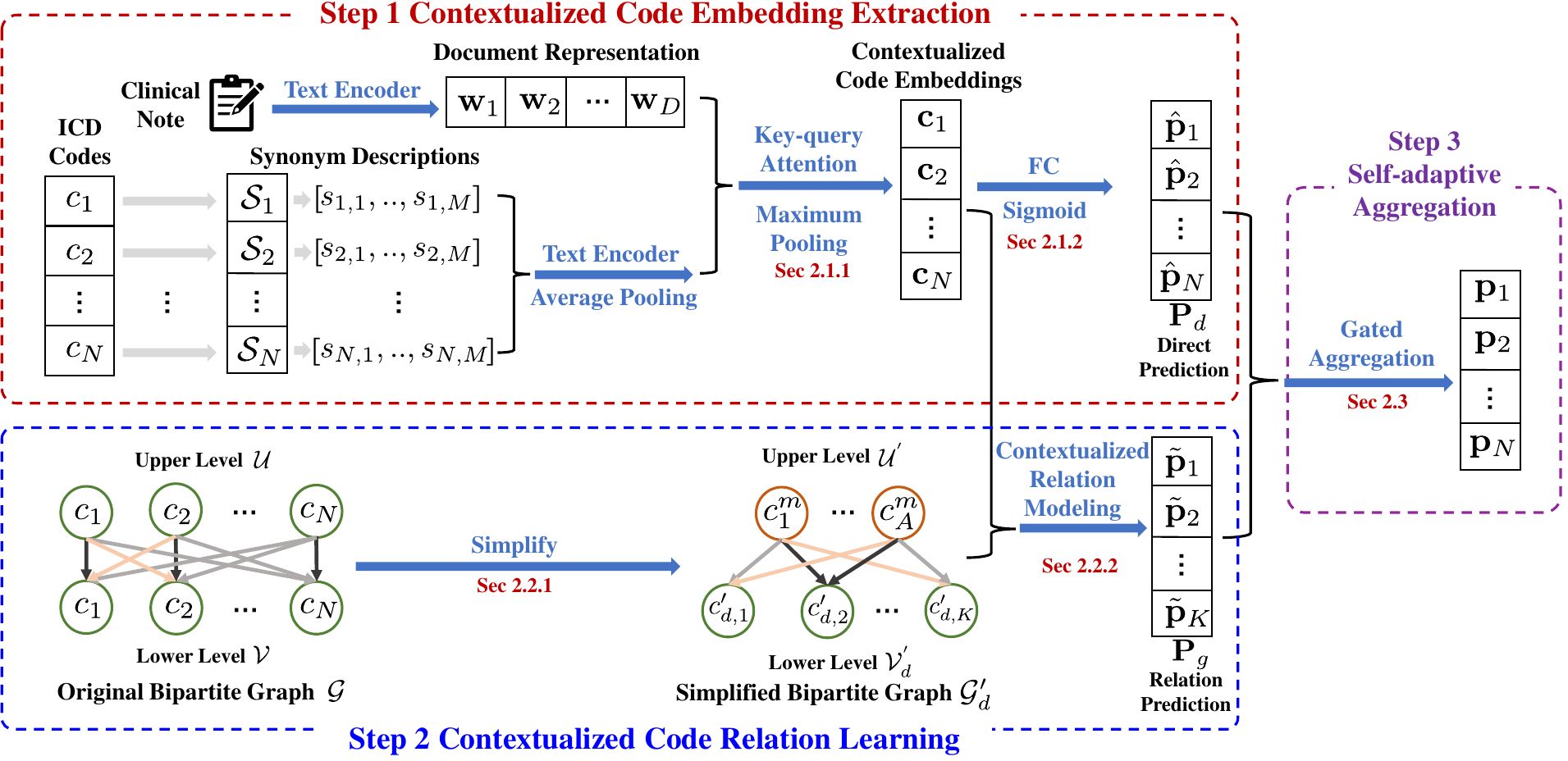}
\vspace{-0.15in}
\caption{The Proposed {\model} structure.}
\vspace{-0.2in}
\label{fig:model}
\end{figure*}
$\bullet$ \textbf{Insufficiently Modeling Relations Among ICD Codes}:
Existing methods typically utilize the ICD code ontology~\cite{xie2019ehr} or the co-occurrence graph~\cite{cao2020hypercore} to model the relationships among ICD codes. However, these approaches only partially capture the code relations. The ontology solely encompasses the ``child-parent'' relation, which aids in enhancing the representation learning of rare ICD codes. On the other hand, the co-occurrence graph can merely indicate whether two codes appear together in the training set's ground truths. These two graphs are insufficiently complex to encompass the intricate relationships among ICD codes. 

For instance, ``696.0'' (\emph{Psoriasis arthropathy}) has been found to have a weak connection with ``579.0'' (\emph{Celiac disease})~\cite{sanchez2018enteropathy}. However, they do not share a common parent code in the ontology and are not connected in the co-occurrence graph, which makes the existing approaches ineffective in modeling this relationship.
Furthermore, both the co-occurrence graph and ICD code ontology fail to capture \emph{exclusive code relations} between different code families, e.g., the relationship between code ``780.60'' (\emph{Fever, unspecified}) and code ``659.2'' (\emph{Maternal pyrexia during labor unspecified})\footnote{http://www.icd9data.com/2014/Volume1/780-799/780-789/780/780.60.htm}. Consequently, there is a need for a novel and effective approach to model the intricate relationships among ICD codes.
%
%
%

$\bullet$ \textbf{Ignoring the Importance of Context}:
The current approaches to ICD coding involve three main steps: (1) ICD code representation learning, (2) clinical note representation learning, and (3) ICD code extraction based on the outputs from the previous two steps. These approaches typically rely on the ICD code ontology or the co-occurrence graph to enhance the representation learning of ICD codes, which is \emph{independent} of the second step. Consequently, the learned code relations remain \textbf{fixed} across all clinical notes.

However, we contend that the context of clinical notes plays a crucial role in ICD coding. For instance, codes belonging to the same sub-category, such as ``488.81'' (\emph{Influenza A with pneumonia}) and ``488.01'' (\emph{Influenza, avian with viral pneumonia}), are generally mutually exclusive since patients typically contract only one type of influenza. Nevertheless, exceptions may arise if a patient contracts both types of influenza simultaneously~\cite{williams2011influenza}, necessitating the assignment of both codes. Therefore, it is imperative to consider the context in order to learn contextualized and dynamic code relations for this task.

\textbf{Our Approach}: 
To overcome these limitations, we propose a novel approach called {\model}\footnote{Source code can be found in the supplementary file \url{https://drive.google.com/file/d/1yv81UffBEIVrQdT-sQYQcbfeXF42176r/view?usp=share_link}.}, which is a contextualized and flexible framework designed to enhance the learning of ICD code representations, as depicted in Figure~\ref{fig:model}. Unlike existing methods that simultaneously learn ICD code and clinical note representations, our approach employs a dependent learning paradigm.

In Step 1, {\model} begins by learning contextualized code embeddings based on the input clinical note $d$ and the multi-synonyms of ICD codes, which follows~\cite{yuan2022code}. These learned code embeddings are then utilized to directly calculate the prediction probabilities $\mathbf{P}_d$ and are subsequently fed into the graph learning phase.

In Step 2, we construct a flexible and contextualized bipartite graph $\mathcal{G} = (\mathcal{U}, \mathcal{V}, \mathcal{E})$ for each clinical note $d$, enabling all codes to communicate with each other through an attention-based strategy. To improve computational efficiency, we propose to reduce the size of nodes in $\mathcal{V}$ by retaining only the top $K$ codes with the highest direct code probabilities estimated in Step 1. Besides, we propose to use coarse-grained ICD code categories to substitute the original codes in $\mathcal{U}$. Most importantly, we use the contextualized code embeddings learned in Step 1 as initializations for the graph update process. 
Thus, the learned relationship is dependent on both the codes and the processed clinical note. The updated code embeddings are then used to calculate the relation code probabilities $\mathbf{P}_g$.

In Step 3, we introduce a self-adaptive gating mechanism that automatically combines the two sets of probabilities, $\mathbf{P}_d$ and $\mathbf{P}_g$, to obtain the aggregated results for the final prediction.
Experimental results on six datasets, including MIMIC-III-50, MIMIC-III-Full, MIMIC-IV-ICD9-50, MIMIC-IV-ICD9-Full, MIMIC-IV-ICD10-50, and MIMIC-IV-ICD10-Full, demonstrate the effectiveness of {\model} compared to state-of-the-art baselines.

In addition, we also propose a novel selective training strategy to reduce the computation cost brought by relation modeling.


\section{Methodology}



The aim of automatic ICD coding is to extract a specific set of ICD codes $\mathcal{C} = \{c_1,\cdots,c_N\}$ from the given clinical note $d=[w_1, \cdots, w_D]$. Here, $N$ and $D$ represent the number of ICD codes and the word count of the clinical notes, respectively. 
Our model, as depicted in Figure~\ref{fig:model}, comprises three major steps:  contextualized code embedding and direct code prediction, contextualized code relation learning, self-adaptive aggregation, and the selective training strategy. In the following subsections, we will provide detailed explanations of each module.

\subsection{Contextualized Code Embedding and Direct Code Prediction}\label{sec:code}
\subsubsection{Contextualized Code Embedding}
Inspired by the recent work MSMN~\cite{yuan2022code}, we propose to use a synonym-based code-wise attention framework to extract contextualized code features from clinical notes by referring to the ICD code synonym description at the initial stage. Specifically, we employ the same text encoder to encode both the raw clinical note and the code descriptions. Thus, we have the clinical note embedding as follows:
\begin{align}
    [\mathbf{w}_1,\cdots, \mathbf{w}_D] = \text{TextEncoder}(d)\label{eq:textencoder},
\end{align}
where $\mathbf{w}_k$ is the representation of the $k$-th word in $d$. 
Each code $c_i\in\mathcal{C}$ contains $M$ synonym descriptions $\mathcal{S}_i=[s_{i,1}, \cdots, s_{i,M}]$, and each $s_{i,j}$ consists of $L$ words. Similarly, for each code synonym, we have $[\mathbf{ws}_1^{i,j}, \cdots, \mathbf{ws}_{L}^{i,j}] = \text{TextEncoder}(s_{i,j})$. We then use the average pool to obtain the overall synonym embedding as follows:
\begin{align}
    \mathbf{s}_{i,j} = \text{Pool}([\mathbf{ws}_1^{i,j}, \cdots, \mathbf{ws}_{L}^{i,j}]).
\end{align}
Next, we utilize a standard key-query attention layer~\cite{vaswani2017attention} to extract contextualized code embeddings from the text embedding sequence as follows:
\begin{equation}
{\mathbf{c}}_{i,j} = \text{KeyQueryAttention}(\mathbf{s}_{i,j}, [\mathbf{w}_1,\cdots, \mathbf{w}_D]),
\label{eq:keyquery}
\end{equation}
where ${\mathbf{c}}_{i,j} $ represents the contextualized code embedding. For each code, we use maximum pooling to obtain an overall representation across all the obtained synonym embeddings:
\begin{equation}
{\mathbf{c}}_i = \text{Pool}([{\mathbf{c}}_{i,1},\cdots, {\mathbf{c}}_{i,M}])
\label{eq:maxpool}.
\end{equation}

\subsubsection{Direct Code Prediction}
Subsequently, we apply a fully connected layer to obtain the prediction weight embedding $\boldsymbol{\alpha}_{i}$ for code $c_i$ from the synonymy code embeddings $[\mathbf{c}_{i,1},\cdots, \mathbf{c}_{i,M}]$ learned by Eq.~\eqref{eq:keyquery}:
\begin{equation}
\boldsymbol{\alpha}_i = \text{FC}_{\alpha}(\text{Pool}([\mathbf{s}_{i,1},\cdots, \mathbf{s}_{i,M}])).
\end{equation}
Finally, the direct code probability is calculated by taking the inner product of the prediction embedding and the extracted contextualized embedding, followed by applying the sigmoid activation function $\sigma()$:
\begin{equation}\label{eq:direct_prob}
\hat{\mathbf{p}}_{i} = \sigma(\boldsymbol{\alpha}_i\cdot{\mathbf{c}}_i).
\end{equation}
$\mathbf{P}_d=[\hat{\mathbf{p}}_{1}, \cdots, \hat{\mathbf{p}}_{N}]$ stands for the initial direct prediction result, and we are going to use contextualized code relation learning to improve it further.

\subsection{Contextualized Code Relation Learning}\label{section:code2code}
To enhance and streamline the ICD coding process, we propose a novel contextualized code relation learning module. This network aims to capture the intricate relationships between ICD codes under the context of the processed case, ultimately improving the accuracy of code assignment.

\subsubsection{Code Relation Graph Construction}\label{section:flex}
\begin{figure}[t]
\centering
\includegraphics[width=0.8\linewidth]{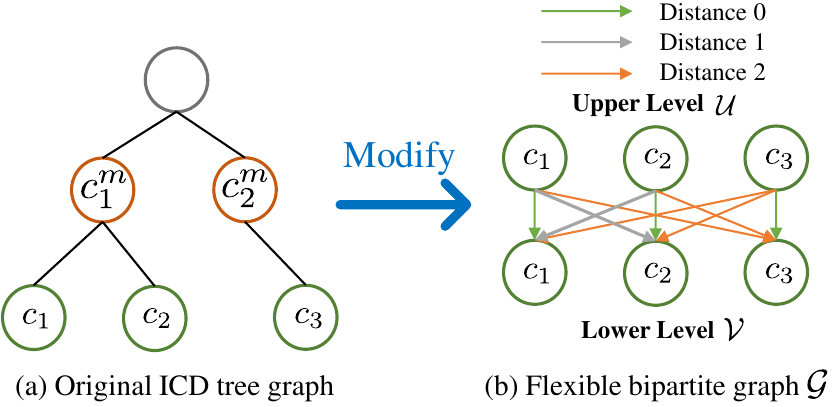}
\caption{We modify the original ICD ontology into a directed flexible bipartite graph. There is an edge for each code pair, and the edge type depends on the distance between two codes on the original ICD ontology.}
\label{fig:bipartite}
\vspace{-0.2in}
\end{figure}
Directly modeling the potential relations on the original ICD ontology is complicated. As illustrated in Figure~\ref{fig:bipartite} (a), $c_1$ can directly exchange information with $c_1^m$. Although there is a path from $c_1$ to $c_3$, the long path decreases the shared information significantly. 
As a result, the model cannot effectively learn the relation between $c_1$ and $c_3$. 

\smallskip
\noindent\textbf{Bipartite Graph Construction}.
To address this issue, we employ a simplified, flexible bipartite graph $\mathcal{G} = (\mathcal{U}, \mathcal{V}, \mathcal{E})$ to represent the code-to-code connections, as depicted in Figure~\ref{fig:bipartite} (b), where $\mathcal{U} = \mathcal{V} = \mathcal{C}$ denote all the ICD codes, and $\mathcal{E}$ denotes the edges. Here, we use the distance (i.e., the number of hops) between any pair of ICD codes on the ontology to represent the edge type.
\emph{Such a design allows each code to communicate with all the codes directly, preserves parts of ICD ontology relations, and fully covers the relations used in the co-occurrence graph.}
The updated embeddings of the lower-level nodes (i.e., $\mathcal{V}$) will be used to calculate relation-enhanced probabilities in the following subsection.

\smallskip
\noindent\textbf{Bipartite Graph Simplification}.
Updating all ICD codes in the constructed graph is time-consuming. In the MIMIC-III-Full setting, the number of targeted ICD codes is 8,922, which means there are a total of $8,922\times 8,922$ edges. To reduce the computational burden, we propose two tricks to reduce the complexity.

\textbf{Lower-level}: \emph{Top-$K$ Code Selection}.
Intuitively, only the codes with higher probabilities calculated by Eq.~\eqref{eq:direct_prob} will be helpful in the final prediction, and the low-probability codes are less likely to affect the result. Thus, we can decrease the size of $\mathcal{V}$ by selecting the top-$K$ codes with the highest probabilities from $\mathbf{P}_d=[\hat{\mathbf{p}}_{1}, \cdots, \hat{\mathbf{p}}_{N}]$, and we use $\mathcal{V}_d^\prime = \{c_{d,1}^\prime, \cdots, c_{d,K}^\prime\}$ denote the reduced lower-level code set, where $c_{d,k}^\prime$ is a selected ICD code. 


\textbf{Upper-level}: \emph{Major Code Substitution}.
In the ICD code ontology, each leaf code belongs to a coarse-grained category. For example, ``250.03 (Diabetes mellitus without mention of complication, type I [juvenile type], uncontrolled)'' belongs to the category ``250 (Diabetes mellitus)'', which is also called the major code. To reduce the size of $\mathcal{U}$, we propose to merge ICD codes belonging to the same coarse-grained category and use the major codes set $\mathcal{U}^\prime=\{c^m_{1}, \cdots, c^m_{A}\}$ as the nodes, where $A$ is the number of the major codes. 


\textbf{Simplified Graph}. 
Given the new nodes $\mathcal{V}_d^\prime$ and $\mathcal{U}^\prime$, we then update the edge set $\mathcal{E}_d$ between $\mathcal{V}_d^\prime$ and $\mathcal{U}^\prime$. Let $\mathcal{E}_d^\prime$ denote the new edges, where each edge type $e_{a,k}^d$ is still the distance between a major code $c_{a}^m$ and a select ICD code $c_{d,k}^\prime$.
In such a way, we can have a simplified bipartite graph $\mathcal{G}_d^\prime = (\mathcal{U}^\prime, \mathcal{V}_d^\prime, \mathcal{E}_d^\prime)$. Note that the set of $\mathcal{V}_d^\prime$ is dependent on the clinical note $d$, but for all clinical notes, the major codes $\mathcal{U}^\prime$ are the same. In such a way, the simplified graph can be considered personalized.

\subsubsection{Contextualized Relation Modeling}\label{sec:cont}
To update the node embeddings on the simplified graph $\mathcal{G}_d^\prime$, we first initialize the node embeddings of both $\mathcal{V}_d^\prime$ and $\mathcal{U}^\prime$. For each selected node $c_{d,k}^\prime \in \mathcal{V}_d^\prime$, we can use the contextualized code embedding learned by Eq.~\eqref{eq:maxpool} as the initialization. For each major code $c_{a}^m \in \mathcal{U}^\prime$, we use Eqs.~(\ref{eq:textencoder}-\ref{eq:maxpool}) to calculate the initialized embeddings. The edge type $e_{a,k}^d$ is randomly initialized as a representation. Therefore, the input of the relation graph is \textbf{customized} for each processed note. 

Next, we use the graph transformer~\cite{dwivedi2020generalization} to model the interaction between codes on the simplified graph $\mathcal{G}_d^\prime$ as follows:
\begin{equation}
    \{\mathbf{U}^*, \mathbf{V}_{d}^*, \mathbf{E}_{d}^*\} = \text{GraphTransformer}(\mathcal{G}_d^\prime).
\end{equation}
The resulting enhanced embeddings $\mathbf{V}_{d}^*=[\tilde{\mathbf{c}}_1, \cdots, \tilde{\mathbf{c}}_K]$ from the lower-level are then used to estimate the relation enhanced code probabilities $\mathbf{P}_g=[\tilde{\mathbf{p}}_{1}, \cdots, \tilde{\mathbf{p}}_{K}]$ as follows:
\begin{equation}
    \tilde{\mathbf{p}}_i = \sigma({\boldsymbol{\beta}}_i\cdot\tilde{\mathbf{c}}_i),
\end{equation}
where ${\boldsymbol{\beta}}_i$ is the prediction weight vector for relation-enhanced code embedding, which is obtained as follows:
\begin{equation}
{\boldsymbol{\beta}}_i = \text{FC}_{\beta}(\text{Pool}([\mathbf{s}_{i,1},\cdots, \mathbf{s}_{i,M}])).
\end{equation}
\subsection{Self-adaptive Aggregation}\label{sec:agg}
To combine the direct prediction probabilities $\mathbf{P}_d=[\hat{\mathbf{p}}_{1}, \cdots, \hat{\mathbf{p}}_{N}]$ and the relation enhanced probabilities $\mathbf{P}_g=[\tilde{\mathbf{p}}_{1}, \cdots, \tilde{\mathbf{p}}_{K}]$, we propose a novel self-adaptive gating module to aggregate the two results.
In detail, we use the raw activation result (element-wise product $\odot$ of $\boldsymbol{\alpha}_{i}$ and ${\mathbf{c}}_i$) from the Section~\ref{sec:code} to calculate the proportion value $\gamma_{i}$ of relation enhanced prediction $\tilde{\mathbf{p}}_{i}$, making the model able to contextually decide the proportion of different inference sources: 
\begin{align}
    \gamma_{i} = \sigma(\text{FC}_{\gamma}(\boldsymbol{\alpha}_i \odot {\mathbf{c}}_i)).
\end{align}
The final prediction result $\mathbf{p}_i$ is the gated combination of two prediction results as follows:
\begin{align}
\mathbf{p}_i = (1-\gamma_{i})\hat{\mathbf{p}}_{i}+\gamma_{i}\tilde{\mathbf{p}}_{i}\label{eq:gate}.
\end{align}
For the codes that are not selected (i.e., not among top $K$), we set the $\gamma_i$ to 0 instead.

The training loss for the prediction results is:
\begin{equation}
     \mathcal{L}_{CE} = \frac{1}{N}\sum_{i=1}^N\text{CrossEntropy}(\mathbf{p}_i, \mathbf{g}_i),
\end{equation}
where $\mathbf{g}_i$ is the ground-truth label.
In the meanwhile, we use the following loss to encourage the model to use less complex relation inference results by making the average proportion value $\gamma_i$ as a loss term as follows:
\begin{equation}
     \mathcal{L}_{comp} =\frac{1}{N}\sum_{i=1}^N \gamma_i.
\end{equation}
The final training loss is:
\begin{equation}
    \mathcal{L} = \mathcal{L}_{CE}+\lambda\mathcal{L}_{comp},
\end{equation}
where $\lambda$ is the hyper-parameters for controlling the importance of ${L}_{comp}$.
\subsection{Selective Training Strategy}\label{app:selective}
In ICD coding, the majority of codes possess a limited number of positive labels. The average positive labels for each case under the MIMIC-III-Full setting is 15. However, there are total of 8,922 labels. Consequently, for full settings, most prediction results consist of negative labels. Despite contributing minimal gradients for model updates, these negative labels still demand equal computational resources during backpropagation. To circumvent this computational inefficiency, we propose a selective training algorithm, detailed in Algorithm~\ref{alg:cap}.
\begin{algorithm}[h]
\small
\caption{Selective Training}\label{alg:cap}
\KwIn{Code set $\mathcal{C}$, dataset $\mathcal{D}$, and model parameter $\theta$.}
\KwOut{Updated model parameter $\theta$.}
\While{Training not finished}{
1. Sampling $d, \text{label set } \mathbf{G}$ from data set $\mathcal{D}$ \\
$(d, \mathbf{G})\gets\mathcal{D}$ \\
2. Perform forward propagation without code relation to estimate the scores.\\
$\hat{\mathbf{P}_{est}} = {\text{\model}}_{wo}(\mathcal{C}, d)$\\
3. Choose top $K_s$ codes.\\
$\mathcal{C}_{K_s}=\{\cdots, c_i, \cdots\}, i \in \text{Top-}K_s(\hat{\mathbf{P}_{est}})$\\
4. Adding additional $K_s$ random codes, and the ground-truth codes to form $\mathcal{C}_{back}$.\\
$\mathcal{C}_{back}=\{\mathcal{C}_{K_s},\mathcal{C}_{ground}, \mathcal{C}_{random}\}$\\
5. Select the labels of $\mathcal{C}_{back}$ from label set $\mathbf{G}$.\\
$\mathbf{G}_{back} = \{\cdots, \mathbf{g}_i, \cdots\}, c_i \in \mathcal{C}_{back}$\\
6. Perform forward calculation for the selected codes only.\\
$\mathbf{P}_{back}= \text{\model}(\mathcal{C}_{back}, d)$\\
7. Calculate the loss $\mathcal{L}$.\\
$\mathcal{L} = LossFunc(\mathbf{P}_{back}, \mathbf{G}_{back})$\\
8. Update model parameter $\theta$.\\
$\theta = \theta- \nabla L$\\
}
\end{algorithm}
Specifically, we initially choose the top $K_s$ training codes based on the output of $\text{\model}_{wo}$ (steps 1-5) and then only perform backpropagation on these chosen codes (steps 6-8). Here, $\text{\model}_{wo}$ means directly predicting the results without using contextualized code relation learning in Section~\ref{section:code2code}. 
This selective training algorithm enables the model to focus on the most relevant codes during training, thereby reducing computational resources and avoiding the expenditure on negative labels that provide minimal gradient information.
\section{Experiments}

\subsection{Experiment Settings}
In this section, we introduce the experimental settings, which include the datasets, baselines, and evaluation metrics. 
\begin{table*}[t]
\centering
\caption{Hypere parameter settings.}
\label{table:para}
\begin{tabular}{l|cccc}
\hline
\multicolumn{1}{c|}{Settings} & Train Epoch & R-Drop factor & $\mathcal{L}_{comp}$ weight $\beta$ & Synonym number $M$ \\ \hline
MIMIC-III-50                  & 40          & 12.5   & 0.001                          & 8                \\
MIMIC-III-Full                & 30          & 5.0    & 0.01                           & 8                \\
MIMIC-IV-ICD9-50              & 15          & 5.0    & 0.001                          & 8                \\
MIMIC-IV-ICD9-Full            & 30          & 5.0    & 0.01                           & 4                \\
MIMIC-IV-ICD10-50             & 15          & 5.0    & 0.001                          & 8                \\
MIMIC-IV-ICD10-Full           & 30          & 5.0    & 0.01                           & 4                \\ \hline
\end{tabular}
\end{table*}
\subsubsection{Implementation Details}\label{app:imp}
We implement our model in PyTorch, training them on an Ubuntu 20.04 system with 128 GB of memory and four NVIDIA A6000 GPUs.

For our model, we employed a single-layer LSTM with a hidden dimension of 512 as the $\text{TextEncoder}$. Word embeddings are initialized using GLOVE pre-trained embeddings on MIMIC-III notes, as described in the work~\cite{vu2021label}.
The attention dimension of the $\text{KeyQueryAttention}$ and $\text{GraphTransformer}$ component is configured to 256. 

The top-$K$ parameter $K$ is set to 300 for all the \textbf{full} settings. For the \textbf{50} settings, $K=50$ is fixed.
For contextualized code relation learning, we use all the major codes as the upper node set $\mathcal{U}^\prime$ of the relation graph for all the settings.

As for the optimizer, we use the Adam optimizer with an initial learning rate of $5e$-$4$, accompanied by linear decay, and early stop is applied by checking the Macro AUC score on the validation set. 
We employed the R-Drop~\cite{wu2021r} regularization technique, as introduced in the previous work MSMN~\cite{yuan2022code}.
Remaining parameters are summarized in Table~\ref{table:para}.

We also apply an efficient selective training strategy in Appendix~\ref{app:selective} to further improve the training speed on the full setting. In short, we only select the top $K_s=1,000$ codes for back propagation.


\subsubsection{Datasets} 
To evaluate our method, we utilize the ICD coding datasets derived from the MIMIC-III~\cite{johnson2016mimic} and MIMIC-IV~\cite{johnson2020mimic} projects. Specifically, we follow the settings of~\cite{mullenbach2018explainable, yuan2022code} to create the MIMIC-III-50 and MIMIC-III-Full datasets, and the settings of~\cite{nguyen2023mimic} to create the MIMIC-IV-ICD9-50, MIMIC-IV-ICD9-Full, MIMIC-IV-ICD10-50, and MIMIC-IV-ICD10-Full datasets. The \textbf{50} setting focuses on evaluating the top 50 most frequent ICD codes, and the \textbf{Full} setting focuses on evaluating all potential ICD codes. 
The statistics of the six datasets are presented in Table~\ref{table:datasets}.

\begin{table}[t]
\centering
\caption{Statistics of the six datasets.}
\label{table:datasets}
\resizebox{0.49\textwidth}{!}{
\tabcolsep=0.05cm
\begin{tabular}{l|cc|cc|cc}
\hline
Database & \multicolumn{2}{c|}{{\textbf{MIMIC-III}}} & \multicolumn{4}{c}{\textbf{MIMIC-IV}}                         \\  \hline
Code Version& \multicolumn{2}{c|}{{ICD9}}                           & \multicolumn{2}{c|}{{ICD9}} & \multicolumn{2}{c}{{ICD10}} \\ \hline
Settings & 50                     & Full                   & 50          & Full       & 50          & Full        \\ \cline{1-7} 
\# of codes in $\mathcal{C}$       & 50                     & 8,922                  & 50          & 11,331     & 50          & 26,096      \\
Training size  & 8,066                  & 47,723                 & 170,664     & 180,553    & 104,077     & 110,442     \\
Validation size    & 1,573                  & 1,631                  & 6,406       & 7,110      & 3,805       & 4,017       \\
Testing  size & 1,729                  & 3,372                  & 12,405      & 13,709     & 7,368       & 7,851       \\ 
Avg \# of tokens   & 1,478                  & 1,434                  & 1,499      & 1,459     &  1,687       &  1,662       \\ \hline
\end{tabular}
}

\end{table}



\subsubsection{Baselines}
We divide baselines into two classes based on whether they utilize pre-trained language models (PLMs) as encoders:
(1) \textbf{Non-PLM} methods include CAML~\cite{mullenbach2018explainable}, MultiResCNN~\cite{li2020icd}, HyperCore~\cite{cao2020hypercore}, LAAT and JointLAAT~\cite{vu2021label}, MSMN~\cite{yuan2022code}, and TwoStage~\cite{nguyen2023two}; and (2)
\textbf{PLM} methods include {KEPT}\footnote{For {KEPT}, we only report the results of the 50 setting since the Full setting results utilize multiple methods to perform a multi-stage retrieval.}~\cite{yang2022knowledge}, {HiLAT}~\cite{liu2022hierarchical}, and {PLM-ICD}~\cite{huang-etal-2022-plm}. 

We obtain the results of baselines either from the released trained models or the original papers if the codes are unavailable. It is worth noting that these models~\cite{liu2022treeman, ng2023modelling, yang2022multi,niu2023retrieve,zhang2022automatic} are not listed in baselines since they require additional information, such as annotation data, information source, and multiple retrieval stages.


\subsubsection{Evaluation Metrics}
Following previous studies~\cite{mullenbach2018explainable, vu2021label, yuan2022code}, we report Macro \& Micro AUC, Macro \& Micro F1, and Precision at K (P@$\mathcal{K}$) metrics, where $\mathcal{K}=5,8, 15$ for different settings. The \textbf{Bold} notation indicates the best results among non-PLM techniques, while \underline{Underline} notation signifies the best results when considering the PLM setting. 

\begin{table}[t]
\centering
\caption{Results on the MIMIC-III-50 test set.}
\label{table:MIMIC-50}
\resizebox{0.49\textwidth}{!}{
\begin{tabular}{l|l|cccccc}
\hline
\multirow{2}{*}{Category}&\multirow{2}{*}{Method} & \multicolumn{2}{c}{AUC}       & \multicolumn{2}{c}{F1}        & \multicolumn{2}{c}{Pre}       \\ \cline{3-8}
     &    & Macro         & Micro         & Macro         & Micro         & P@5           & P@8           \\ \hline
\multirow{3}{*}{{PLM}}
&{HiLAT}                     & 92.7          & 95.0          & 69.0          & {\underline{73.5}}    & 68.1          & 55.4          \\
&{PLM-ICD}                   & 90.2          & 92.7          & 64.8          & 69.6          & 65.0          & 53.0          \\
&{KEPT}                      & 92.6          & 94.8          & 68.9          & 72.9          & 67.3          & 54.8          \\ \hline

\multirow{8}{*}{\makecell[l]{Non-PLM}} 
&CAML                      & 87.5          & 90.9          & 53.2          & 61.4          & 60.9          & -             \\
&MultiResCNN               & 89.9          & 92.8          & 60.6          & 67.0          & 64.1          & -             \\
&HyperCore                 & 89.5          & 92.9          & 60.9          & 66.3          & 63.2          & -             \\
&LAAT                      & 92.5          & 94.6          & 66.6          & 71.5          & 67.5          & 54.7          \\
&JointLAAT                 & 92.5          & 94.6          & 66.1          & 71.6          & 67.1          & 54.6          \\
&TwoStage                  & 92.6          & 94.5          & 68.9          & 71.8          & 66.7          & -             \\
&MSMN                      & 92.8          & 94.7          & 68.3          & 72.5          & 68.0          & 54.8          \\

&\cellcolor{blue!25}{\model}                & \cellcolor{blue!25}\textbf{93.3} & \cellcolor{blue!25}\textbf{95.1} & \cellcolor{blue!25}\textbf{69.3} & \cellcolor{blue!25}\textbf{73.1} & \cellcolor{blue!25}\textbf{68.3} & \cellcolor{blue!25}\textbf{55.6} \\ \hline

\end{tabular}
}

\end{table}

\begin{table*}[h]
\centering
\caption{Results on the MIMIC-IV-50 test sets.}
\label{table:MIMIC-IV-50}
\resizebox{1\textwidth}{!}{
\begin{tabular}{l|l|ccccc|ccccc}
\hline
\multirow{3}{*}{Category}& \multirow{3}{*}{Method} &\multicolumn{5}{c|}{MIMIC-IV-\textbf{ICD9}-50}  & \multicolumn{5}{c}{MIMIC-IV-\textbf{ICD10}-50} \\ \cline{3-12}                                                                                                 
& & \multicolumn{2}{c}{AUC}                                               & \multicolumn{2}{c}{F1}                                                & Pre  & \multicolumn{2}{c}{AUC}                                               & \multicolumn{2}{c}{F1}                                                & Pre                              \\ \cline{3-12} 
    &                   & Macro                             & Micro                             & Macro                             & Micro                             & P@5   & Macro                             & Micro                             & Macro                             & Micro                             & P@5                             \\ \hline

PLM & {PLM-ICD}                                    & 95.0                              & 96.4                              & 71.4                              & 75.5                              & 62.4                              & 93.4                              & 95.6                              & 69.0                              & 73.3                              & 64.6                              \\\hline
\multirow{5}{*}{Non-PLM} 
& CAML                                       & 93.1                              & 94.1                              & 65.3                              & 69.2                              & 58.6   & 91.1                              & 93.2                              & 64.3                              & 67.6                              & 59.6                              \\
& LAAT                                       & 94.9                              & 96.3                              & 70.0                              & 74.5                              & 62.0                              & 93.2                              & 95.5                              & 68.2                              & 72.6                              & 64.4                              \\
& JointLAAT                                  & 94.9                              & 96.3                              & 69.9                              & 74.3                              & 62.0                              & 93.4                              & 95.6                              & 68.4                              &  72.9                        & 64.5                              \\
& MSMN                                       & 95.1                              & 95.5                              & 71.9                              & 75.8                              & 62.6                              & 93.6                              & 95.7                              & 70.3                              & 74.2                              & 65.2                              \\
\rowcolor{blue!25} 
\cellcolor{white}& {\model}                       & \textbf{95.4}                     & \textbf{96.7}                     & \textbf{72.5}                     & \textbf{76.0}                     & \textbf{62.9}                     & \textbf{93.8} & \textbf{96.0} & \textbf{70.6} & \textbf{74.4} & \textbf{65.4} \\ \hline

\end{tabular}
}
\end{table*}

\subsection{Results of the 50 Setting}
In this section, we present the experimental outcomes for the three \textbf{50} settings, emphasizing the top 50 codes. In this setting, the size of $\mathcal{V}_d^\prime = 50$ for all data. A comprehensive comparison of methodologies on the MIMIC-III and MIMIC-IV settings is provided in Table~\ref{table:MIMIC-50} and Table~\ref{table:MIMIC-IV-50} respectively.
Results reveal that our proposed model, {\model}, outperforms all non-PLM methods across all metrics within the 50 settings. Even when compared to PLM methods, our method still demonstrates state-of-the-art performance on every metric, except for Micro F1 on the MIMIC-III-50 setting. Consequently, our method's advantage in the 50 settings is considerably substantial.

\begin{table}[ht]
\centering
\caption{Results on the MIMIC-III-Full test set.}
\label{table:MIMIC-FULL}
\resizebox{0.49\textwidth}{!}{
\begin{tabular}{l|ccccccc}
\hline
\multirow{2}{*}{Method} & \multicolumn{2}{c}{AUC}       & \multicolumn{2}{c}{F1}        & \multicolumn{3}{c}{Pre}                       \\ \cline{2-8} 
                      & Macro         & Micro         & Macro         & Micro         & P@5           & P@8           & P@15          \\ \hline
{PLM-ICD}                                    & 92.5          & 98.9          & 8.4           & 58.0          & \underline{83.9}    & \underline{76.7}    & \underline{61.1}    \\ \hline
CAML                                       & 89.5          & 98.6          & 8.8           & 53.9          & -             & 70.9          & 56.1          \\
MultiResCNN                                & 91.0          & 98.6          & 8.5           & 55.2          & -             & 73.4          & 58.4          \\
HyperCore                                  & 93.0          & 98.9          & 9.0           & 55.1          & -             & 72.2          & 57.9          \\
LAAT                                       & 91.9          & 98.8          & 9.9           & 57.5          & 81.3          & 73.8          & 59.1          \\
JointLAAT                                  & 92.1          & 98.8          & \textbf{10.7} & 57.5          & 80.6          & 73.5          & 59.0          \\
TwoStage                                   & 94.6          & 99.0          & 10.5          & 58.4          & -             & 74.4          & -             \\
MSMN                                       & 95.0          & \textbf{99.2} & 10.3          & 58.4          & 82.5          & 75.2          & 59.9          \\
\rowcolor{blue!25} 
{\model}                                & \textbf{95.2} & \textbf{99.2} & 10.2          & \textbf{59.1} & \textbf{83.4} & \textbf{76.2} & \textbf{60.7} \\ \hline
\end{tabular}
}
\end{table}

\begin{table*}[h]
\centering
\caption{Results on the MIMIC-IV-Full test sets.}
\label{table:MIMIC-IV-FULL}
\resizebox{1\textwidth}{!}{
\begin{tabular}{l|l|ccccc|ccccc}
\hline
\multirow{3}{*}{Category}& \multirow{3}{*}{Method} &\multicolumn{5}{c|}{MIMIC-IV-\textbf{ICD9}-Full}  & \multicolumn{5}{c}{MIMIC-IV-\textbf{ICD10}-Full} \\ \cline{3-12}                                                                                                 
& & \multicolumn{2}{c}{AUC}                                               & \multicolumn{2}{c}{F1}                                                & Pre  & \multicolumn{2}{c}{AUC}                                               & \multicolumn{2}{c}{F1}                                                & Pre                              \\ \cline{3-12} 
    &                   & Macro                             & Micro                             & Macro                             & Micro                             & P@8   & Macro                             & Micro                             & Macro                             & Micro                             & P@8                             \\ \hline

PLM & {PLM-ICD}                                   & 96.6       & 99.5       & 14.4       & \underline{62.5}      & \underline{70.3}        & 91.9      & 99.0      & 4.9       & 57.0     & 69.5                              \\\hline
\multirow{5}{*}{Non-PLM} 
& CAML                                       & 93.5       & 99.3       & 11.1       & 57.3      & 64.9        & 89.9      & 98.8      & 4.1       & 52.7     & 64.4\\
& LAAT                                      & 95.2       & 99.5       & 13.1       & 60.3      & 67.5              & 93.0      & 99.1      & 4.5       & 55.4     & 67.0  \\
& JointLAAT                                  & 95.6       & 99.5       & 14.2       & 60.4      & 67.5     & 93.6      & 99.3      & 5.7       & 55.9     & 66.9 \\
& MSMN                                       & \textbf{96.8}       &  \textbf{99.6}       & 13.9       & 61.2      & 68.9     & 97.1      & \textbf{99.6}      & 5.4       & 55.9     & 67.7  \\
\rowcolor{blue!25} 
\cellcolor{white}& {\model}                  & \textbf{96.8}       &  99.5       & \textbf{15.0}       & \textbf{62.4}      & \textbf{70.1}      &\textbf{97.2}       & \textbf{99.6}       & \textbf{6.3}        & \textbf{57.8}      & \textbf{70.0}\\ \hline

\end{tabular}
}
\end{table*}

\subsection{Results of the Full Setting}
Next, we examine the results of three \textbf{Full} settings in Table~\ref{table:MIMIC-FULL} and Table~\ref{table:MIMIC-IV-FULL}.
For the Full settings, only one PLM method is included since other PLM methods cannot handle the huge potential label space as illustrated in Table~\ref{table:datasets}.
{\model} once again outperforms existing non-PLM methods across a majority of metrics within the MIMIC-III-Full, MIMIC-IV-ICD9-Full, and MIMIC-IV-ICD10-Full contexts. 
Concurrently, it is important to note that the improvement of {\model} in Precision at $\mathcal{K}$ (P@$\mathcal{K}$) scores is particularly significant, demonstrating that the proposed relation learning technique is effective in refining high-probability code predictions.
Compared to PLM methods, although PLM-ICD exhibits superior performance in some P@$\mathcal{K}$ metrics, its underwhelming performance across other metrics - including its poor results in the 50 settings - renders it less competitive. Besides, our model operates with considerably fewer parameters (22 Million vs. 120 Million) compared to PLM methods. As such, the advantage of the {\model} remains evident.

\subsection{Ablation Study}
In this section, we present an ablation study to examine the contribution of each proposed module to the overall performance of our model. The results are summarized in Table~\ref{table:ABA}. The following notations represent different configurations of our model:

\begin{itemize}
\item \textbf{W/O Relation}: This notation signifies our proposed model without the application of the relation learning discussed in Section~\ref{section:code2code}.

\item \textbf{W/O Flexible Relation Graph (W/O FRG)}: This is the configuration where our proposed flexible relation graph, described in Section~\ref{section:flex}, is replaced with a fixed ICD ontology + co-occurrence graph, similar to~\cite{cao2020hypercore}.

\item \textbf{W/O Context}: This denotes a version of our model that excludes the use of contextualized code embeddings, as per Section~\ref{sec:cont}. The non-contextualized code description embeddings $\mathbf{s}_{i,j}$ are employed to initialize the relation graph $\mathcal{G}_d^\prime$, similar to~\cite{xie2019ehr, cao2020hypercore}.

\item \textbf{W/O Self-Adaptive Aggregation (W/O SAA)}: This indicates our proposed model without the integration of the self-adaptive aggregation discussed in Section~\ref{sec:agg}. The $\mathbf{P}_g$ is directly employed as the final output.
\end{itemize}

We initiate the comparison with \textbf{W/O Relation} and the proposed model. Under both the 50 and Full settings, the omission of the relation results in a decline in most metrics, signifying the efficacy of our proposed relation learning in managing both frequent code settings and full code settings. Concurrently, we individually substitute components of the proposed contextualized relation learning to evaluate the efficiency of each proposed module. \textbf{W/O FRG} incurs the smallest decrease, yet it does not suggest that the contribution of flexible relation modeling is trivial. As the flexible relation graph centers on intricate and weak relationships, its enhancements are less apparent in quantitative analysis. The case studies in Section~\ref{sec:case} will illustrate that the relations inferred within the flexible graph hold significant value.
Finally, both \textbf{W/O SAA} and \textbf{W/O Context} contribute to substantial declines in the final performance. This underlines the significance of contextualized relation modeling and self-adaptive aggregation. 
In conclusion, all the proposed module contribute to the ultimate performance.
\begin{table*}[h]
\centering
\caption{Results of ablation experiments on the MIMIC-III datasets.}
\label{table:ABA}
\resizebox{1\textwidth}{!}{
\begin{tabular}{ccccccc|ccccccc}
\hline
Dataset & \multicolumn{6}{|c|}{MIMIC-III-50}                                                                                                                                                                                                                     & \multicolumn{6}{c}{MIMIC-III-Full}                                                                                                                                                                  \\ \hline
\multicolumn{1}{l|}{\multirow{2}{*}{Method}}   & \multicolumn{2}{c}{AUC}                                         & \multicolumn{2}{c}{F1}                                          & \multicolumn{2}{c|}{Pre}                                         & \multicolumn{2}{c}{AUC}                                         & \multicolumn{2}{c}{F1}                                          & \multicolumn{3}{c}{Pre}                                         \\\cline{2-14}
\multicolumn{1}{c|}{}                          & Macro                          & Micro                          & Macro                          & Micro                          & P@5                            & P@8                            & Macro                          & Micro                          & Macro                          & Micro                          & P@5                            & P@8      & P@15                       \\ \hline
\rowcolor{blue!25}
\multicolumn{1}{l|}{\model} & \textbf{93.3} & \textbf{95.1} & \textbf{69.3} & \textbf{73.1} & \textbf{68.3} & \textbf{55.6} & \textbf{95.2} & \textbf{99.2} & 10.2  & \textbf{59.1} & \textbf{83.4} & \textbf{76.2} & \textbf{60.7}\\
\multicolumn{1}{l|}{W/O Relation}              & 93.1                           & 95.0                           & 69.0                           & 72.6                           & 68.1                           & 55.2                           & 95.2                           & 99.1                           & 9.3                            & 58.9                           & 82.8                           & 75.7     & 60.5                      \\ \hline
\multicolumn{1}{l|}{W/O FRG}              & 93.2                           & 95.1                           & 69.0                           & 72.9                           & 68.2                           & 55.5                           & 95.1                           & 99.2                           & 10.0                           & 58.8                           & 83.3                           & 76.0             & 60.5              \\
\multicolumn{1}{l|}{W/O Context}               & 92.0                           & 93.7                           & 66.4                           & 70.0                           & 66.2                           & 53.8                           & 95.0                           & 99.1                           & \textbf{10.7} & 57.9                           & 81.4                           & 74.3               & 59.4            \\
\multicolumn{1}{l|}{W/O SAA}              & 92.5                           & 94.7                           & 68.6                           & 72.2                           & 67.9                           & 55.0                           & 95.0                           & 99.1                           & 9.7                            & 58.8                           & 82.9                           & 75.9        & 60.1                   \\ \hline
\end{tabular}
}
\end{table*}

\subsection{Case Study on Learned Code Relations}\label{sec:case}
\begin{figure}[t]
\centering
\includegraphics[width=0.9\linewidth]{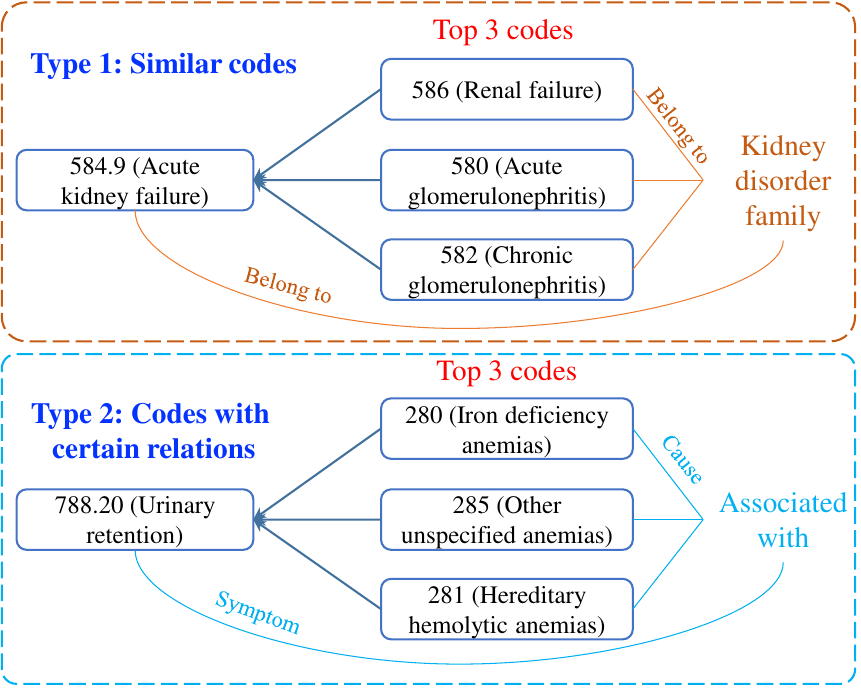}
\caption{Two typical learned code relation cases.}
\label{fig:case}
\end{figure}

To better comprehend the efficacy of our proposed flexible, context-aware code relationship learning in facilitating code prediction, we conduct case studies to elucidate the relationships inferred from the code. The cases are depicted in Figure~\ref{fig:case}. Based on our analysis, the inferred relationships can be categorized into two types.

\textbf{Type 1} relationships concentrate on differentiating similar codes by referencing codes that originate from the same family. For example, the top-3 referenced codes for the code ``584.9'' (\textit{Acute kidney failure}) are ``586'' (\textit{Renal failure}), ``580'' (\textit{Acute glomerulonephritis}), and ``582'' (\textit{Chronic glomerulonephritis}). All these referenced codes, including ``584.9'', are part of the \textit{Kidney disorder family}. The model, by taking these analogous codes into consideration, can make more precise predictions by discerning subtle differences.

\textbf{Type 2} relationships focus on referencing codes that demonstrate specific associations. For instance, the top-3 cited codes for ``788.20'' (\textit{Urinary retention}) are ``280'' (\textit{Iron deficiency anemias}), ``286'' (\textit{Other unspecified anemias}), and ``281'' (\textit{Hereditary hemolytic anemias}), all falling under the \textit{Anemia} category. The incidence of \textit{Urinary Retention} is frequently associated with the severity of \textit{anemia}~\cite{hung2015association}. By utilizing flexible relationship learning, the model is capable of recognizing such associations and employing them to bolster prediction outcomes.

To summarize, the relationships inferred within the flexible graph are highly interpretable and exhibit a robust correlation with real-world medical practices.

\subsection{Evaluation of Top-$K$ Code Selection}

\begin{table}[t]
\centering
\footnotesize
\caption{Evaluation of $K$ values on the MIMIC-III-Full.}
\label{table:k}
\begin{tabular}{l|cc|cc|c}
\hline
\multirow{2}{*}{$K$} & \multicolumn{2}{c|}{AUC} & \multicolumn{2}{c|}{Pre} & \multirow{2}{*}{Memory} \\ \cline{2-5}
                           & Macro       & Micro      & P@5        & P@8        &                         \\ \hline
$300$                    &  95.2     &  99.2    & 83.4       & 76.2       & 9.15 GB                 \\
$200$                    &  95.2     &  99.2    & 83.5       & 76.1       & 7.78 GB                 \\
$100$                    &  95.2     &  99.2    & 83.1       & 75.8       & 6.41 GB                 \\
$50$                     &  95.1     &  99.2    & 83.0       & 75.4       & 5.05 GB                 \\ \hline
\end{tabular}
\end{table}
In this section, we evaluate the efficacy of the top-$K$ code selection strategy, delineated in Section~\ref{section:flex}. We manipulate $K$ within a range of 50 to 300, and the corresponding performance metrics along with per-sample GPU memory costs are compiled in Table~\ref{table:k}. It is clear that escalating values of $K$ bolster the performance of our model, albeit at the cost of increased GPU memory consumption. Nevertheless, the marginal performance enhancement decelerates as $K$ keeps surging, predominantly due to the dwindling influence of the remaining low-probability codes, elucidated in Section~\ref{section:flex}. This indicates that our proposed top-$K$ code selection strategy is potent, without compromising the overall model performance.

\subsection{Evaluation of Major Code Substitution}
\begin{table}[t]
\centering
\footnotesize
\caption{Evaluation of Major Code Substitution on the MIMIC-III-50 dataset.}
\label{table:major}
\resizebox{1\linewidth}{!}{
\begin{tabular}{l|cc|c|c|c}
\hline
\multirow{2}{*}{Method}       & \multicolumn{2}{c|}{AUC} & \multirow{2}{*}{P@5} & \multirow{2}{*}{Memory} & \multirow{2}{*}{Time} \\ \cline{2-3}
                                & Macro      & Micro      &                      &                         &                       \\ \hline
\multicolumn{1}{c|}{Complete}  & 93.3       & 95.1       & 68.4                 & 6.0 GB                  & 0.23s                  \\
\multicolumn{1}{c|}{Major} & 93.3       & 95.1       & 68.3                 & 1.5 GB                  & 0.05s                 \\ \hline
\end{tabular}}

\end{table}
We also assess the efficacy of the proposed major code substitution strategy in Section~\ref{section:flex}, and the results are shown in Table~\ref{table:major} on the MIMIC-III-50 dataset. 
The term \textbf{Complete} indicates the strategy where the upper-level codes, denoted as $\mathcal{U}$, are not replaced with the major codes $\mathcal{U}^\prime$. Conversely, \textbf{Major} refers to the implementation of the major code substitution strategy. The reported speed and memory costs are the average per-sample metrics observed during training.
As can be deduced from Table~\ref{table:major}, employing the major code substitution results in a substantial decrease in memory usage and computational time while sustaining a performance level comparable to the complete approach.

\subsection{Evaluation of the Selective Training}
\begin{table}[t]
\centering
\caption{GPU memory space comparison results on the MIMIC-III-Full setting.}
\label{table:Eff}
\begin{tabular}{c|cc}
\hline
     Method                                                        & Memory  \\ \hline
{\model} Selective & 9.15 GB                   \\
{\model} W/O Selective                                                       & 72.76GB                  \\ \hline
\end{tabular}
\end{table}
To prove the efficiency of the proposed selective training method, we report the per-sample space cost results in Table~\ref{table:Eff}.
From Table~\ref{table:Eff}, we can discover that our proposed selective training approach significantly reduces memory cost. By utilizing a more efficient training strategy that selectively samples the code space, our method drastically reduces GPU memory usage from 72.76 GB to just 9.15 GB by nearly 12.5\% of the original cost.

\section{Related Work}
The goal of automatic ICD coding is to infer and assign ICD codes based on the textual clinical note. Currently, a majority of automatic ICD coding techniques, such as CAML~\cite{mullenbach2018explainable} and MultiResCNN~\cite{li2020icd}, employ a dual framework for ICD code prediction. In particular, clinical notes and codes are independently converted into embeddings. Then, a code-wise attention framework is utilized to extract relevant information from the encoded clinical notes based on code embeddings. Further enhancements to this approach are proposed from multiple perspectives.

Some studies involve leveraging supplementary information or knowledge, for instance, using ICD code descriptions to initialize ICD code embeddings~\cite{dong2021explainable,zhou2021automatic}. MSMN~\cite{yuan2022code} expands on this by incorporating synonym descriptions of ICD codes. 
Additionally, there have been works to improve code representations using ICD relation data. MSATT-KG~\cite{xie2019ehr} and \citeauthor{teng2020explainable} utilize ICD ontology to enrich the initial code embeddings, while HyperCore~\cite{cao2020hypercore} employs an extra co-occurrence graph to enhance code embeddings. However, as outlined in Section~\ref{sec:intro}, those methods fall short in effectively modeling code relationships. Other research like LAAT~\cite{vu2021label} and TwoStage~\cite{nguyen2023two}  propose to predict codes in a hierarchical manner to optimize the final prediction outcomes. Concurrently, numerous studies~\cite{liu2022treeman, ng2023modelling, zhang2022automatic, wang2020study}  explore more complex methodologies for incorporating external knowledge. Nonetheless, despite these enhancements, these techniques exhibit limited flexibility when dealing with various ICD coding settings due to their dependence on supplementary information sources or resources. Besides the knowledge, there are also studies~\cite{yang2022multi, niu2023retrieve} that recommend the use of multiple-stage retrieval methods for performance enhancement.

The advent of pre-trained language models (PLMs) has inspired many works to leverage PLMs to enhance ICD coding performance~\cite{huang2022plm, michalopoulos2022icdbigbird, ng2023modelling, kang2023automatic}. However, these methods encounter drawbacks due to the substantial computational cost and the over-fitting problem of PLM models. Furthermore, these PLM-based methods frequently under-perform when compared to simpler baseline models, such as LSTM and CNN~\cite{ji2021does, pascual2021towards}. Despite these drawbacks, certain approaches, such as KEPT~\cite{yang2022knowledge} and HiLAT~\cite{liu2022hierarchical}, have succeeded in markedly improving the performance of PLM-based methods. They achieve this through the application of prompt-based prediction and hierarchical encoding methods. Nevertheless, these approaches still grapple with the issue of high computational costs.

\section{Conclusion}
In this paper, we propose a novel contextualized code relation-enhanced ICD coding model. The proposed model, referred to as {\model}, aims to model the complex yet contextualized relations among ICD codes. {\model} delivers state-of-the-art performance in comparison to current advanced ICD coding systems on six ICD coding datasets, yet it does so while consuming fewer computational resources without using pre-trained language models.
Furthermore, we have undertaken an exploration of the learned code relation within our proposed method. The evidence suggests that our approach is also highly explainable.


\bibliography{anthology,custom}
\bibliographystyle{acl_natbib}



\end{document}